\documentclass[twoside,twocolumn]{article}

\usepackage{blindtext} % Package to generate dummy text throughout this template 

\usepackage[sc]{mathpazo} % Use the Palatino font
\usepackage[T1]{fontenc} % Use 8-bit encoding that has 256 glyphs
\usepackage{microtype} % Slightly tweak font spacing for aesthetics

\usepackage[english]{babel} % Language hyphenation and typographical rules

\usepackage[hmarginratio=1:1,top=22mm,bottom=22mm, left=25mm, right=25mm,columnsep=20pt]{geometry} % Document margins
\usepackage[hang, small,labelfont=bf,up,textfont=it,up]{caption} % Custom captions under/above floats in tables or figures
\usepackage{booktabs} % Horizontal rules in tables

\usepackage{lettrine} % The lettrine is the first enlarged letter at the beginning of the text

\usepackage{enumitem} % Customized lists
\setlist[itemize]{noitemsep} % Make itemize lists more compact

\usepackage{abstract} % Allows abstract customization
\usepackage{titlesec} % Allows customization of titles
\renewcommand\thesection{\Roman{section}} % Roman numerals for the sections
\renewcommand\thesubsection{\roman{subsection}} % roman numerals for subsections
\titleformat{\section}[block]{\large\scshape\centering}{\thesection.}{1em}{} % Change the look of the section titles
\titleformat{\subsection}[block]{\large}{\thesubsection.}{1em}{} % Change the look of the section titles

\usepackage{fancyhdr} % Headers and footers
\usepackage{titling} % Customizing the title section

\usepackage{hyperref} % For hyperlinks in the PDF

\usepackage{algorithm}
\usepackage[noend]{algpseudocode}
\usepackage{amsmath}
\makeatletter
\def\BState{\State\hskip-\ALG@thistlm}
\makeatother
\usepackage{graphicx}
\usepackage{subcaption}

\pretitle{\begin{center}\Large\bfseries} % Article title formatting
\posttitle{\end{center}} % Article title closing formatting
\title{Improved Algorithm for Seamlessly Creating Infinite Loops from a
Video Clip, while Preserving Variety in Textures} % Article title
\author{%
\textsc{Kunjal Panchal}\\[1ex] % Your name
\normalsize University of Massachusetts, Amherst \\ % Your institution
\normalsize \href{mailto:kpanchal@umass.edu}{kpanchal@umass.edu} % Your email address
}
\date{\today} % Leave empty to omit a date
\renewcommand{\maketitlehookd}{%
\begin{abstract}
\par 
This project implements the paper ``Video Textures'' by Szeliski \cite{article}. The aim is to create a ``Moving Picture'' or as we popularly call it, a GIF; which is ``somewhere between a photograph and a video''. The idea is to input a video which has some repeated motion (the texture), such as a flag waving, rain, or a candle flame. The output is a new video that infinitely extends the original video in a seamless way. In practice, the output isn't really infinte, but is instead looped using a video player and is sufficiently long as to appear to never repeat.
\par
Our goal from this implementation was to: improve distance metric by switching from a crude sum of squared distance to most sophisticated wavelet-based distance; add intensity normalization, cross-fading and morphing to the suggested basic algorithm. We also experiment on the trade-off between variety and smoothness.
\end{abstract}
}

\begin{document}

% Print the title
\maketitle

%----------------------------------------------------------------------------------------
%	ARTICLE CONTENTS
%----------------------------------------------------------------------------------------

\section{Introduction}

\lettrine[nindent=0em,lines=3]{I}n current age of technology, GIFs[Graphics Interchange Format] are
popular because of their relatively smaller size compared to other formats. Moreover, loading images online is quicker without losing its quality. And they simply convey the message better than a static image. In the paper “Video Textures” by Schodl, Szeliski, and Essa \cite{article}; they created a `'new type of medium'' called a video texture, which is a continuous infinitely varying stream of images, same concept as a GIF. They presented techniques for analyzing a video clip to extract its structure, and for synthesizing a new, similar looking video of arbitrary length,
which is seamless. The goal is that the method needs only few frames out of the whole clip, it works infinitely continuously with varying patterns to give an illusion of arbitrarily long video clip. We implement this method from scratch.\par
But some clips like water, fire, grass in the wind, candle flames, flags face some jitter (conspicuous jumps between frames) - paper mentions this can solved by cross-fading, but they have not implemented it, we will, in this project, try to remove all the jitters; this is easily fixed by modifying Distance Matrix D so that the pairwise distance between frames also considers a few frames around it. This can be achieved by filtering D using a length 2 or 4 filter with weights set to the binomial coefficients (to approximate a Gaussian distribution, with the correct
width). Next, in video clips featuring time lapses, the lighting changes too drastically when we choose only few frames from many. This can be solved by some pre-processing – normalizing intensities.\par
Then, there are some future work suggestions in the end: use a better distance metric, better blending and maintaining variety. We implement Chebyshev Distance and Wavelet-based Distance; compare the results. Then, we use morphing for better blending, that can do away with jitters. For maintaining variety, we define a parameter which controls randomness. It penalizes a lack of variety in the generated sequences.\par
Such a parameter would enforce that most (if not all) of the frames of the given input sequence are sometimes played and probabilistically vary the generated order of frames.

%------------------------------------------------

\section{Related Background Work}
Our main reference for the base algorithm is the original paper on Video Textures \cite{article} where they called a GIF, a video texture depicting a certain repeatable pattern and their extensions which include the display of dynamic scenes on webpages, the creation of dynamic backdrops for special effects and games, and the interactive control of video-based animation. One of our main enhancements are to create time-lapses using the same base algorithm, we looked for ways to normalize the color intensity of a video as a pre-processing step. The paper on ``Efficient fluorescence image normalization for time lapse movies'' \cite{colornorm} provides great insight where the concerned lighting is fluorescent, which infers a time-dependent back-ground signal and the image gain without the use of additional fluorescent substances. They first tiled the full image into small sub-images and determined background tiles by clustering the statistical moments of the individual intensity distributions. For each image, they interpolated the full background from the identified tiles and thus reconstituted the time-dependent background image. Second, they estimated the time-independent image gain from the background tiles of all pixels and all time points.\par
A thesis on ``Probabilistic Time Lapse Video'' \cite{colornorm1} provides basic to advanced histogram techniques for color normalization in images. The thesis also talked about reduction of noise, jitter removal and normalization of color levels methods being employed in the preprocessing stage to clean the images and provide a suitable starting point for the implementation of the time lapse video creation algorithms.\par
Later, we explored techniques of distance measurements other than sum of squared distance. We first used wavelet transforms to compress each frame into its second level representation. The paper on ``A Study on Wavelet Compression Images Based on Global Threshold'' \cite{wavelet} analyzes the combination between different wavelet filters to select one that give the best compression ratio. In that work they proposed a new type of global threshold to improve the wavelet compression technique. The aim was to maintain the retained energy and to increase the compression ratio. The work on ``Real-time compression and decompression of wavelet-compressed images'' \cite{real} talks about compressing images stored as collections of tiled line textures representing breadth-first trees and then the image is decompressed directly on a GPU employing a microcode pixel shader.\par
To cluster the compressed images, there is lot of literature available on online k-means algorithms \cite{k1, k2, k3, k4}.

%------------------------------------------------

\section{Approach and Algorithm Analysis}
There are two main parts to the algorithm. First, the frames, which are unique enough to create a GIF that will include most of the variety found in the video clip, must be extracted from the input clip. Second, a new clip/ GIF that synthesizes the separate frames must be created.\par
But, the essential concept behind the algorithm is this: given a frame of a clip, we can select a plausible next frame by picking a similar frame, which might be similar enough to be in continuity of the previous frame, but not so similar that we don't get motion or variety in the GIF at all; to the one that would have been played in the original video clip. This next frame may not be the actual next frame in the input, but it may be. In this way, we can infinitely and smoothly extend a video.

\subsection{Extracting the Video Patterns}
To extract the video patterns, we need to compute how much alike the pairs of frames are to each other. This can be achieved by calculating the sum of squared difference between each pair of frames, and storing the results in a distance matrix, $D$. (More sophisticated ways are discussed in the upcoming sections). \par
From there, we can compute a probability matrix, $P$, which assigns probabilities between pairs of frames. $P$ can then be used to calculate the next frame in the
output video given a current frame: since a frame is a row, we discretely sample the next frame using the probability distribution across a row of $P$. The pseudo-code is below:
\begin{algorithm}
\caption{Extracting the Next Frame}\label{nextframe}
\begin{algorithmic}[1]
\Procedure{ExtractFrames}{}
\State \text{\# Construct D, the distance matrix}
\State $D \gets \text{pairwise distance between all frames,}$
\State\qquad ${using}\textit{SSD}$
\State $\text{shift D to the right by \emph{1},} $
\State $\text{to align the next frame \emph{with} potential new}$
\State $\text{frames}$
\State \text{\# Construct D, the distance matrix}
\State $\sigma \gets \text{average of non-zero } D \text{ values }$
\State \qquad$ \ast \text{ SIGMA\_MULTIPLE}$.
\State $P \gets \exp(-D/\sigma)$.
\State $\text{Normalize P so that the}\text{ sum of row \emph{is} \textbf{1} }$
\EndProcedure
\end{algorithmic}
\end{algorithm}
\par
We discuss how to set the SIGMA\_MULTIPLE correctly in the next sections.

\subsection{Preserving Dynamics}
\begin{figure}[!h]
  \centering
  \includegraphics[width=\linewidth]{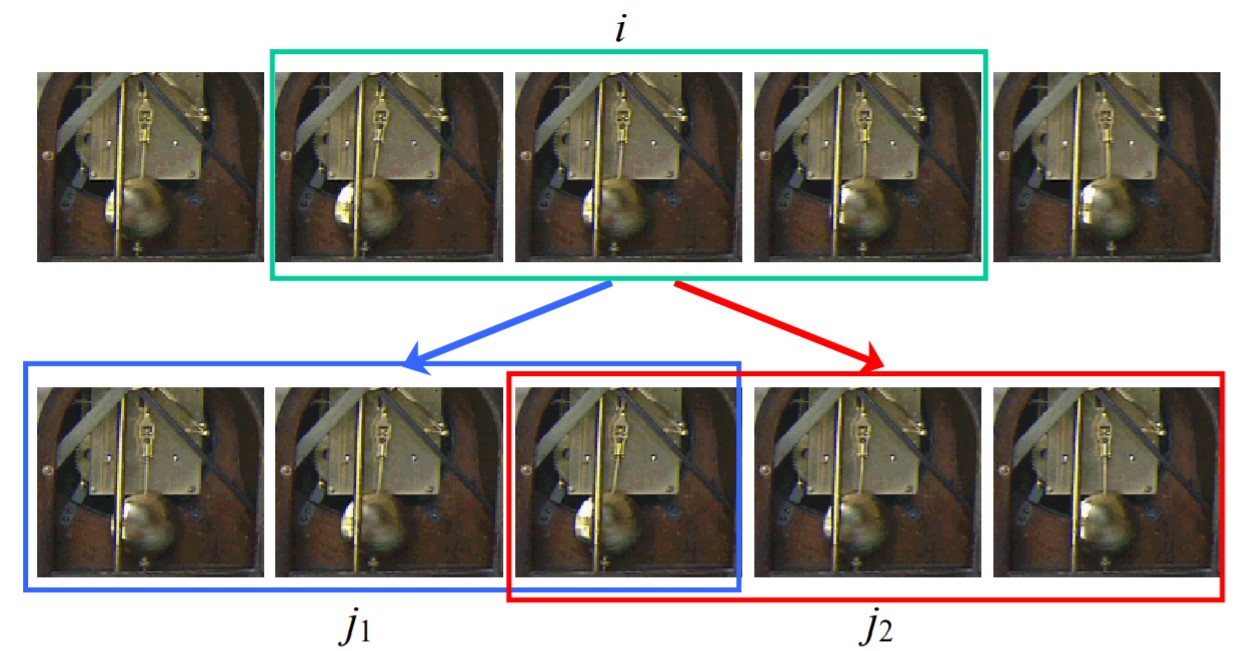}
  \caption{Effect and Need of Dynamics Preservation. A GIF of a pendulum should preserve the direction in which it is moving}
  \label{fig:pend}
\end{figure}

In some cases, the input video has a fluid motion that the GIF should preserve. The paper \cite{article} gives the example of a pendulum swinging in Figure \ref{fig:pend}: the algorithm described in \ref{nextframe} doesn't account for the fact that original video has a side-to-side motion. 
So, the resulting texture may jitter back and forth since there's no distinction that the next frame may have come from the left side or the right side in the original video. \par
This is easily fixed by modifying $D$ so that the pairwise distance between frames also considers a few frames around it. This can be achieved by filtering $D$ using a length 2 or 4 filter with weights set to the binomial coefficients (to approximate a Gaussian distribution, with the correct width). The pseudo-code for this modification is below:

\begin{algorithm}
\caption{Weighted Probability Associated with Each Frame}\label{weightframe}
\begin{algorithmic}[1]
\Procedure{WeightFrames}{}
\State \text{\# Modify D, to preserve motion}
\State $w \gets \text{a 2 or 4 length filter}$
\State\qquad $\text{with }\textit{binomial weights}$
\State $w \gets diag(w)$
\State $\text{filter D with w}$
\State $\text{crop D along the edges due to the filter}$
\State $\text{\# From here, continue on with making P as}$
\State $\text{described above}$
\EndProcedure
\end{algorithmic}
\end{algorithm}

\noindent
The impact on D can be visualized below in Figure \ref{fig:D}:

\begin{figure}[!h]
  \centering
  \includegraphics[width=\linewidth]{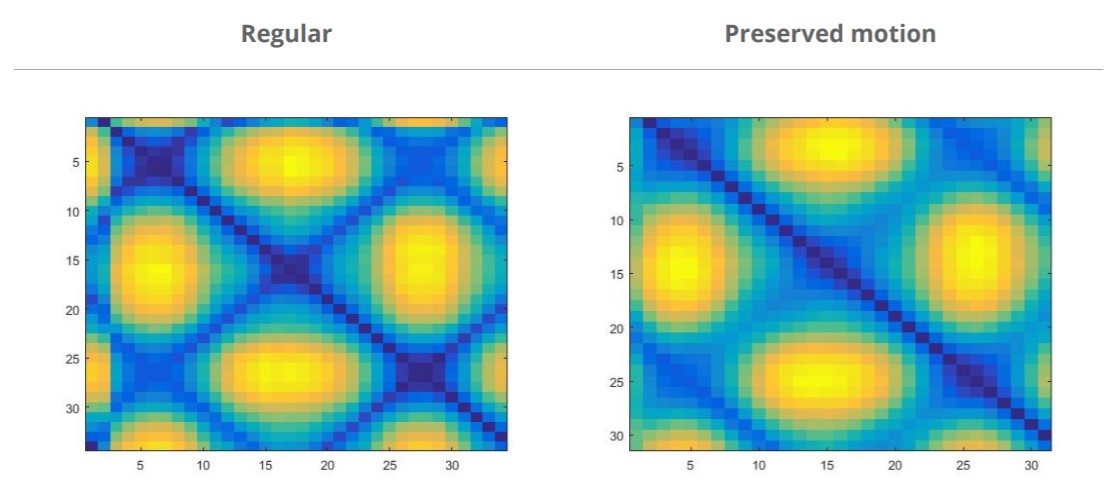}
  \caption{Motion Preservation: Blurring can be observed along the diagonals, and $D$ is slightly smaller in overall size due to the cropping}
  \label{fig:D}
\end{figure}

%------------------------------------------------

\section{Writing the Video Loops}
Once suitable transitions have been identified, we can then generate the GIF. The
original paper \cite{article} describes two approaches to do so: random playback, and video loops. Random playback was just sticking any two frames which were close enough distance-wise, and it didn't provide much logic or sophistication for our main goal of jitter removal. So here, we just discuss a more logical method, which gave us satisfying results.\par
The motivation behind this method is to preselect a sequence of loops that can be repeated, so that the resulting clip smoothly repeats when played on a conventional video player's repeat setting. We attempted to implement dynamic programming algorithm and scheduling algorithm. We had to prune the input video clip, because in many cases, the very start and the end of the video clip are very different from the actual content of the clip.\par
In dynamic algorithm, we have a collection of potential frames; we know that the first and the last frame must be the one and the same. So we take a potential first frame, and go backward towards to same frame by using the concepts from ``Longest Common Subsequence'', where a subsequence is a sequence that can be derived from another sequence by deleting some elements without changing the order of the remaining elements. Longest common subsequence (LCS) of 2 sequences is a subsequence, with maximal length, which is common to both the sequences.\par
Here, we just change the ``characters'' with ``frames'' and find a longest sequence. The reason we want it to be longest is because we need to have as many unique frames in the GIF as possible, to preserve the variety of the original clip.

%------------------------------------------------

\section{Methods}
Till now we talked about the basic algorithm with few improvements here and there, here I will focus on mainly the 5 significant improvements I have implemented to the original methodology. Those 5 ideas are as follows:
\begin{enumerate}
\item Smoothening the Jitters: Cross-Fading
\item Smoothening the Jitters: Blending through Morphing
\item For drastic changes in light intensities: Normalizing
\item Better Distance Metric: Wavelet-transform based Clustering
\item Variety Preservation: Finding the correct hyper-parameters
\end{enumerate}

\subsection{Smoothening the Jitters with Cross-Fading}
Many times, we will find that there doesn't exist a smooth transition between two frames. Which will result in conspicuous jitters. To avoid that, we might be lax in our choice of potential next frames, but that might make our loop/ GIF too short or with much lesser number of unique frames, which won't capture the essence of the clip.\par
One solution is to prefer variety over smoothness, and remove jitters by cross-fading each transition between frames.\par
The trick is to not show frame $i$ for $t$ milliseconds and then switch to frame $i+1$ while completely ``turning off'' the frame $i$; but we have to gradually decrease the opacity (alpha channel) of frame $i$, while we have started showing frame $i+1$, that too, by gradually \textit{increasing} the opacity or the alpha value. \par
\begin{figure}[h!]
 \centering
 \begin{subfigure}[b]{0.48\linewidth}
 \includegraphics[width=\linewidth]{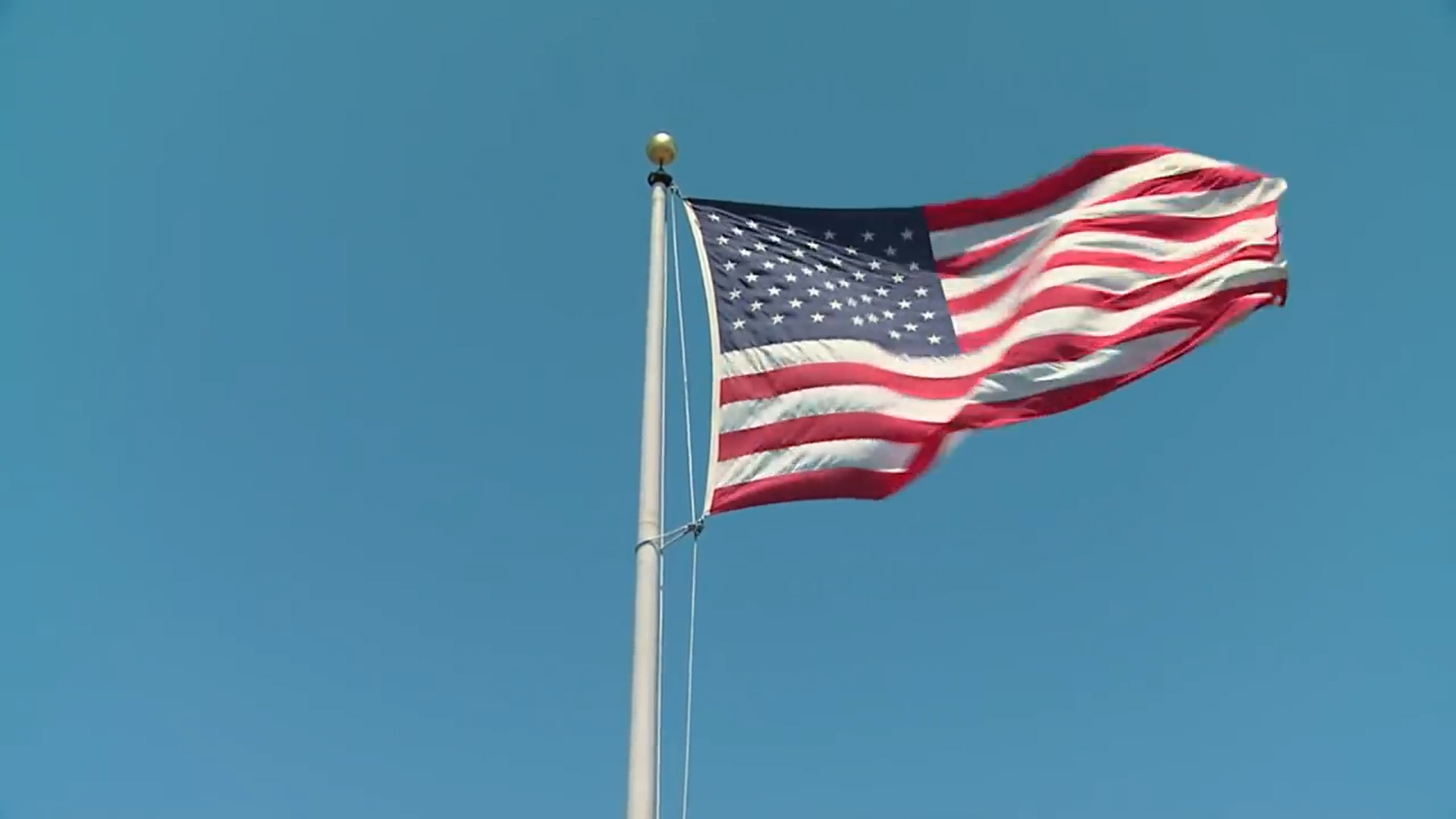}
 \caption{Frame $i$}
 \end{subfigure}  
 \begin{subfigure}[b]{0.48\linewidth}
 \includegraphics[width=\linewidth]{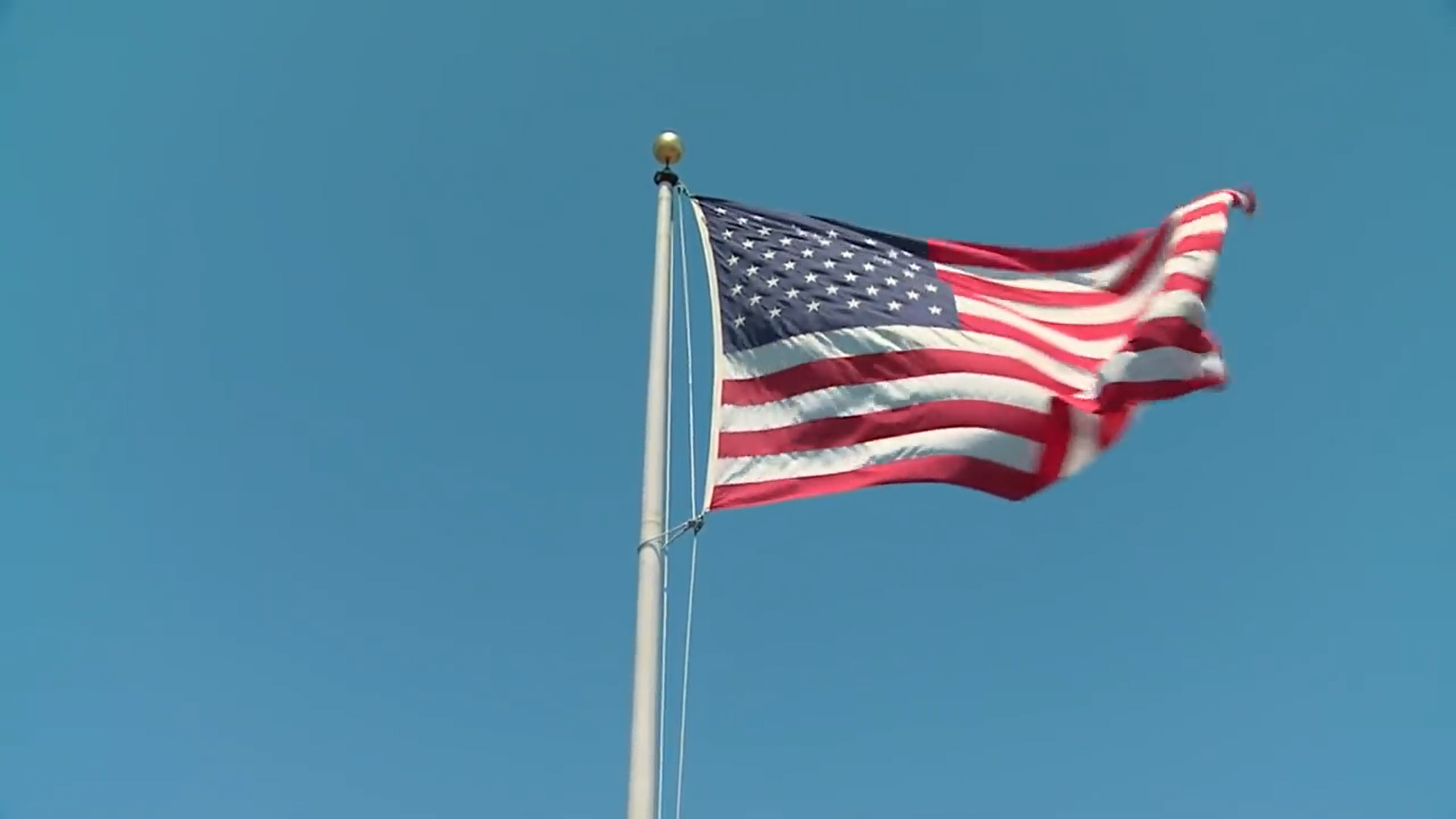}
 \caption{Frame $i+1$}
 \end{subfigure}
 \begin{subfigure}[b]{\linewidth}
 \includegraphics[width=\linewidth]{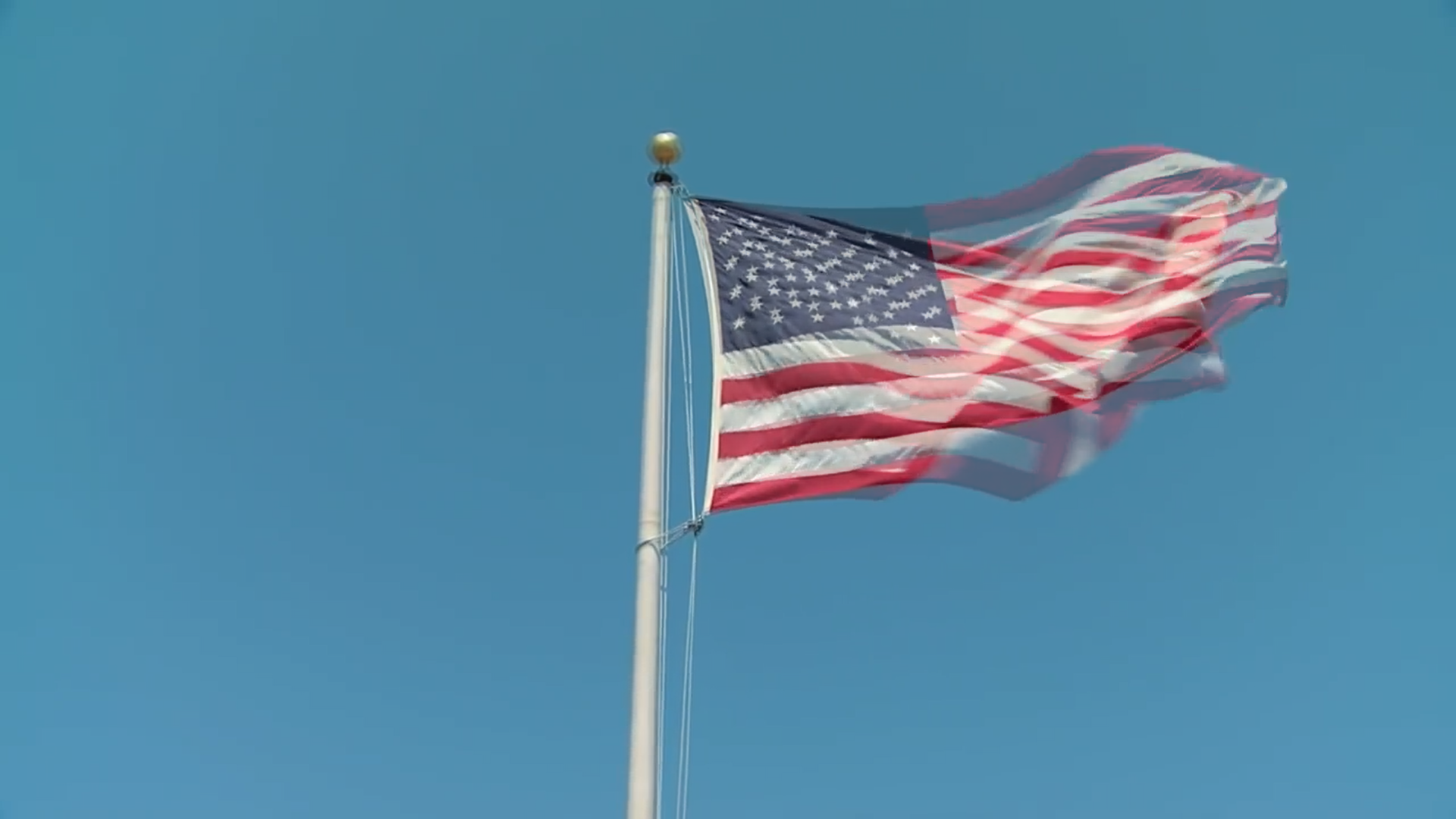}
 \caption{Transition between Frame $i$ and $i+1$}
 \end{subfigure}
 \caption{Example of Cross-Fading where we decrease alpha value of Frame $i$ as we increase alpha value of $i+1$}
 \label{fig:fade}
\end{figure}

That way that jitter between the two frames will not be as apparent as it previously was. Figure \ref{fig:fade} shows the results of cross-fading between two frames in a clip where a flag in waving in the wind.

\subsection{Smoothening the Jitters with Morphing}
A more advanced technique would be to implement morphing, as it would actually transform one frame to the next one, instead of just making the transitional jitters less conspicuous.\par
We can warp Frame $i$ features to Frame $i+1$ features by homography transforms. We first extracted SIFT features from Frames $i$ and $i+1$.\par
\begin{figure}[!h]
  \centering
  \includegraphics[width=\linewidth]{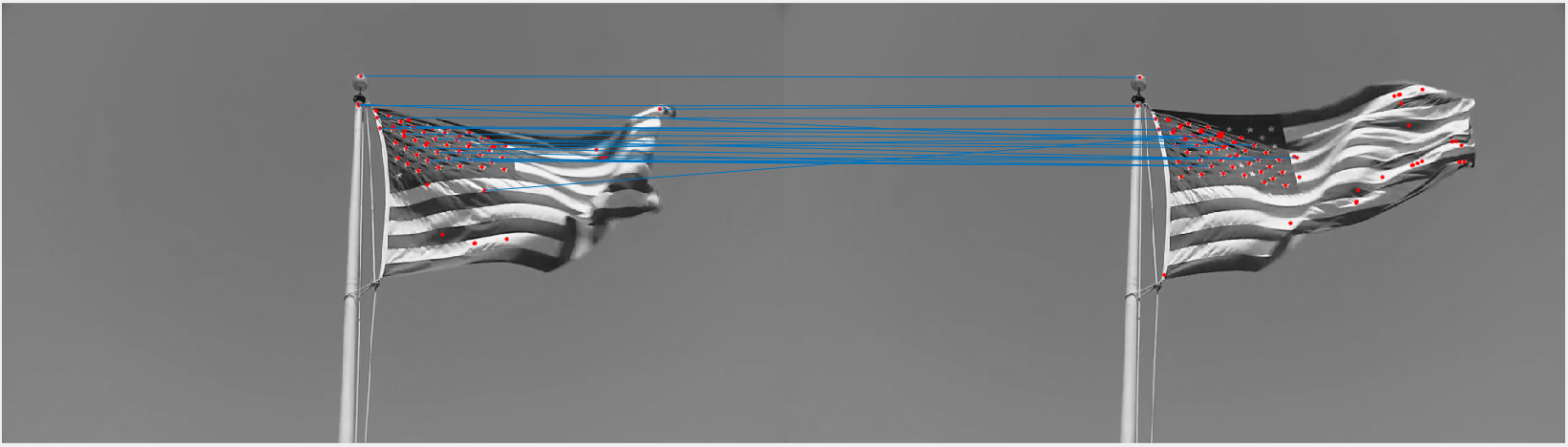}
  \caption{Feature Extraction of source and destination frame, we can warp the geometry of those features in a way that results in smoother transition by inserting the intermediate transition frames}
  \label{fig:SIFT}
\end{figure}

The SIFT features are extracted as shown in Figure \ref{fig:SIFT}.
\par 
The algorithm for interpolating intermediate frames as given in 

\begin{algorithm}
\caption{Interpolation of Intermediate Frames}\label{intermediate}
\begin{algorithmic}[1]
\Procedure{IntermediateFrames}{}
\BState \emph{for}: r = \text{each row of Frame} $i$
\State \emph{for}: c = \text{each column of Frame} $i+1$
\State $q \gets Frame1[r, c]$
\State $p \gets Frame2[r, c]$
\State \text{\# The intermediate FrameN is a combination}
\State \text{\# of the Frame1 and Frame2}
\State $FrameN[r, c] \gets p + (1 - i/n)\ast(q - p)$
\State \emph{end}
\BState \emph{end}
\BState \text{\# Repeat this N times}
\BState \text{\# If N is larger, smoother transition}
\BState \text{\;\;but larger sized GIF}
\BState \text{\# If N is smaller, less smoother transition} 
\BState \text{\;\;but smaller sized GIF}
\EndProcedure
\end{algorithmic}
\end{algorithm}

\subsection{Normalizing the Light Intensity}
To create a GIf of timelapses of a city, the GIF will start with sunny sky and end with night sky, and must be linear so there aren't drastic line changes in the flow.\par
To avoid the sudden changes with respect to sky color, we can first normalize the colors of whole video clip and then apply the same algorithm to extract and synthesize the frames. 
\par ``Efficient fluorescence image normalization for time lapse movies'' \cite{colornorm} talks about the same topic but focuses on fluorescent colors. A theses from University of Edinburgh, ``Probabilistic Time Lapse Video'' \cite{colornorm1} expands the idea of Histogram based contrast stretching and Histogram Equalization and apply a color transformation as follows:
\[
\text{Frame}'_{i(r, c)} = 
\begin{cases}
0, & \text{Frame}_{i(r, c)} \leq V_{min}  \\
255\cdot \frac{\text{Frame}_{i(r, c)} - V_{min}}{V_{max} - V_{min}}, & \parbox[t]{.05\textwidth}{$V_{min} < \text{Frame}_{i(r, c)} < V_{max}$}\\
255, & \text{Frame}_{i(r, c)} \geq V_{min}
\end{cases}
\]
where ${Frame}_{i(r, c)}$ and ${Frame}'_{i(r, c)}$ are frames before and after the intensity stretching respectively; $V_{min}$ and  $V_{max}$ are the color values equivalent to the upper and lower bounds of the distribution 1\% and 99\% \cite{colornorm1}.

\subsection{Distance Measurement with Wavelet based Clustering}
I tried to explore better distance metric than sum of squared distance and focused on binning very similar images together. \par
Using a wavelet transform, the wavelet compression methods are adequate for representing transients, such as high-frequency components in two-dimensional images. This means that the transient elements of a data signal can be represented by a smaller amount of information than would be the case if some other transform, such as the more widespread discrete cosine transform, had been used.\par
So idea here to just get the high level details of each frame and cluster all the similar images in one cluster, while increasing the interclass distance and decreasing the intraclass distance. After using wavelet transforms, we use K-means algorithm. It can be interpreted as a greedy algorithm for approximately minimizing a loss function related to data compression.\par
\begin{figure}[!h]
  \centering
  \includegraphics[width=\linewidth]{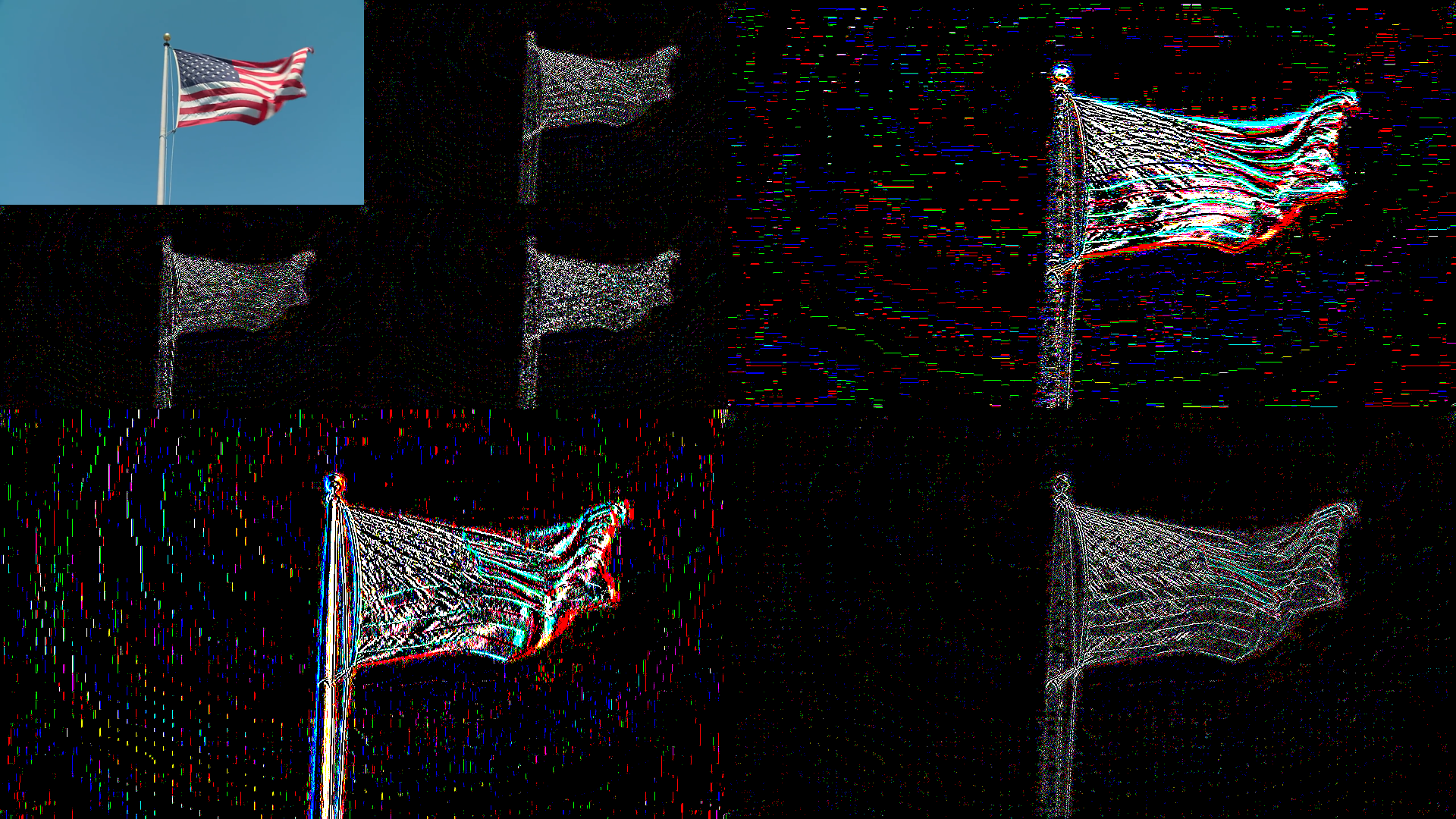}
  \caption{A two-level 2D discrete wavelet transform}
  \label{fig:wav}
\end{figure}

Figure \ref{fig:wav} shows a two level compression of a single frame from our flag clip.
\par
After compressing all the frames, we can run K-means algorithm. The choice of $K$ is arbitrary but keeping the computational resources in mind and size of the output GIF in check, I have decided to keep $K$ between 1 to 10. I ran K-means for $K = 1$ to $10$, and visualized and quantified the clusters to pick the optimal $K$ manually which minimizes the intraclass distance and maximizes the interclass distance.\par
Now that the clusters are decided, we need to choose a strategy to pick frames which will play in continuity.\par
One strategy is to pick one initial frame, then pick the next one from the nearest cluster which is not the same cluster as the current frame came from. And the next to next frame from the cluster which was not previous or previous to previous. That way we will get variety and smooth transition as the next frame will be nearer to the current but not so near that it won't induce the variety.
\par 
Paper ``A Study on Wavelet Compression Images Based on Global Threshold'' \cite{wavelet} gives a good insight into this.

\subsection{Hyper-parameters}
The probability matrix $P$ is created using an exponential function and dividing by a constant, $\sigma$. The paper \cite{article} notes that $\sigma$ is (often, but in our case always) set to a small multiple of the average non-zero $D$ values. The user-set parameter SIGMA\_MULTIPLE controls $\sigma$ further: smaller values of $\sigma$ force the best transitions to be taken, and larger values allow for more randomness. We typically set SIGMA\_MULTIPLE to 0.05.

%Text requiring further explanation\footnote{Example footnote}.

%------------------------------------------------

\section{Results}

\begin{figure}[!h]
  \centering
  \includegraphics[width=\linewidth]{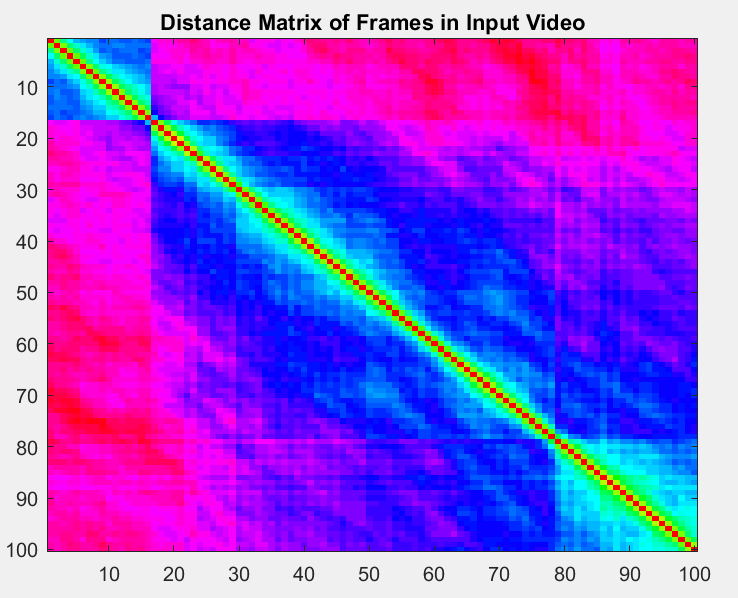}
  \caption{Distance Matrix of Frames in Input Video}
  \label{fig:1}
\end{figure}

First we show the results of using SSD distance metric. Figure \ref{fig:1} show raw distance between Frame $i$ and each Frame $j$ where $j \in \{1, \dots, \#\text{NumberOfFrames}\}$

\par
\begin{figure}[!h]
  \centering
  \includegraphics[width=\linewidth]{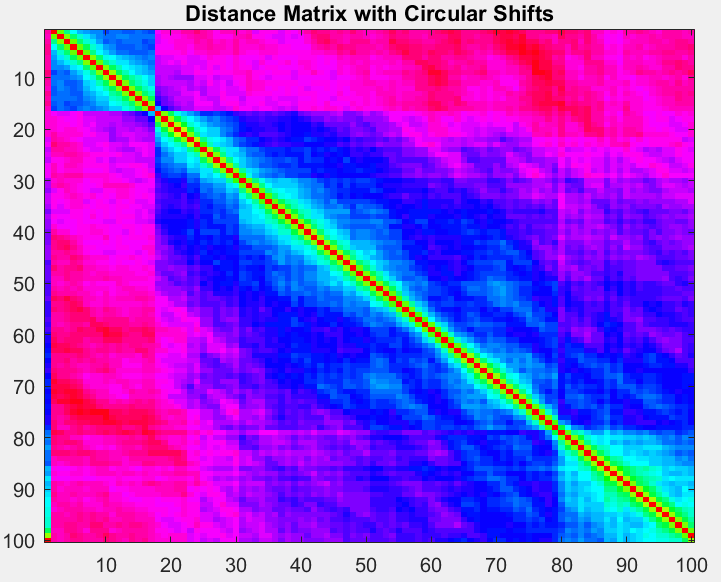}
  \caption{Distance Matrix with Circular Shifts}
  \label{fig:2}
\end{figure}

We must remember that we need to make a seamless loop out of the video clip, so we shift the distance matrix circularly to get the rows with minimum difference between  the first cell and the last cell. Figure \ref{fig:2} shows the circular shift on the distance matrix.

\par
\begin{figure}[!h]
  \centering
  \includegraphics[width=\linewidth]{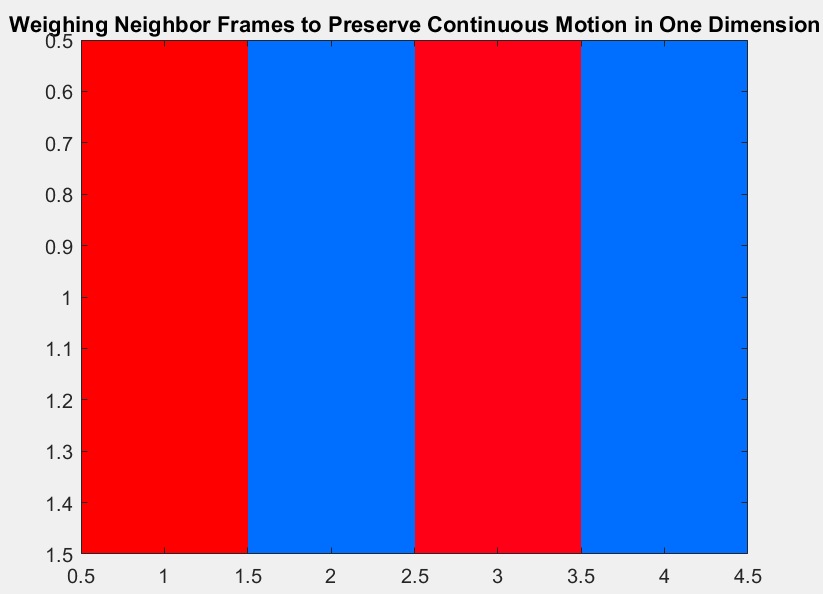}
  \caption{Weighted Neighbor Frames to Preserve Continuous Motion in One Direction}
  \label{fig:3}
\end{figure}
\begin{figure}[!h]
  \centering
  \includegraphics[width=\linewidth]{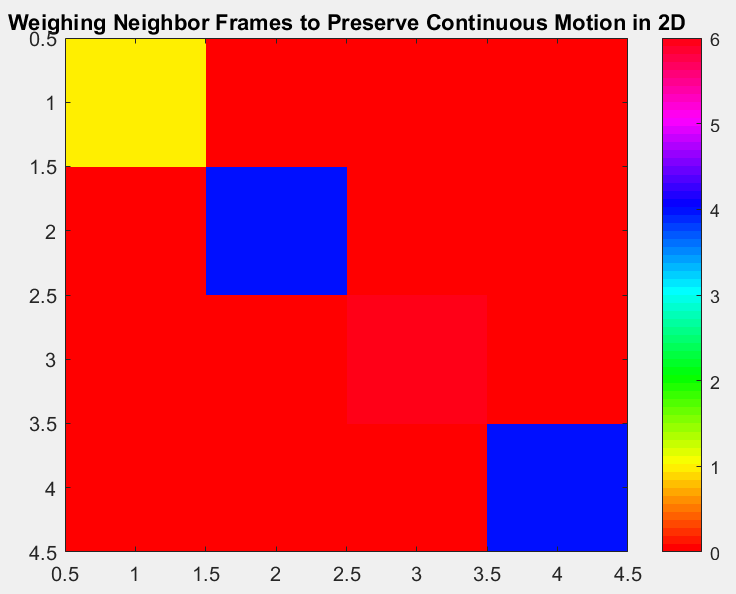}
  \caption{Weighted Neighbor Frames to Preserve Continuous Motion in Two Direction}
  \label{fig:4}
\end{figure}

Now we calculate the weights of neighbor frames so we can assign probability of potential next frame to preserve continuous motion. Figure \ref{fig:3} shows four weighted neighbors in one dimension and Figure \ref{fig:4} shows the same in two dimensions.

\par
\begin{figure}[!h]
  \centering
  \includegraphics[width=\linewidth]{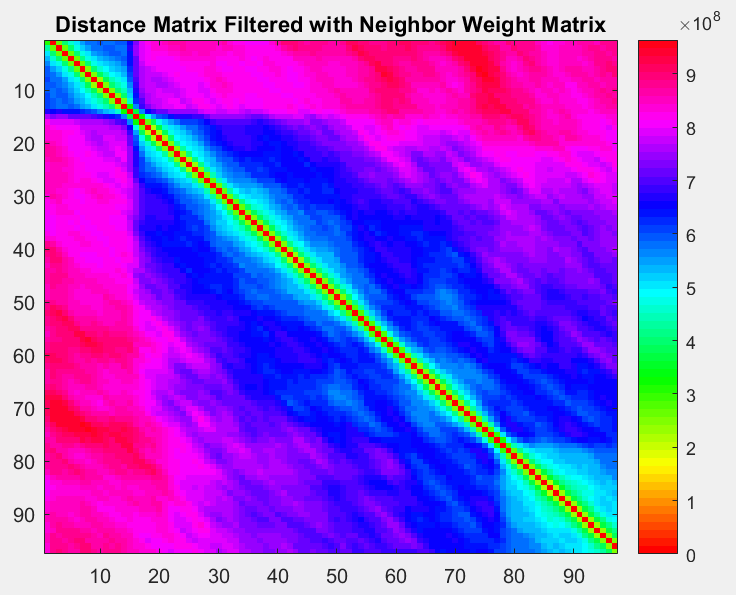}
  \caption{Distance Matrix with Neighbor Weight Matrix}
  \label{fig:5}
\end{figure}

We use $P = \exp(-D/\sigma)$ to get the weighted distance matrix as shown in Figure \ref{fig:5}.

\par
\begin{figure}[!h]
  \centering
  \includegraphics[width=\linewidth]{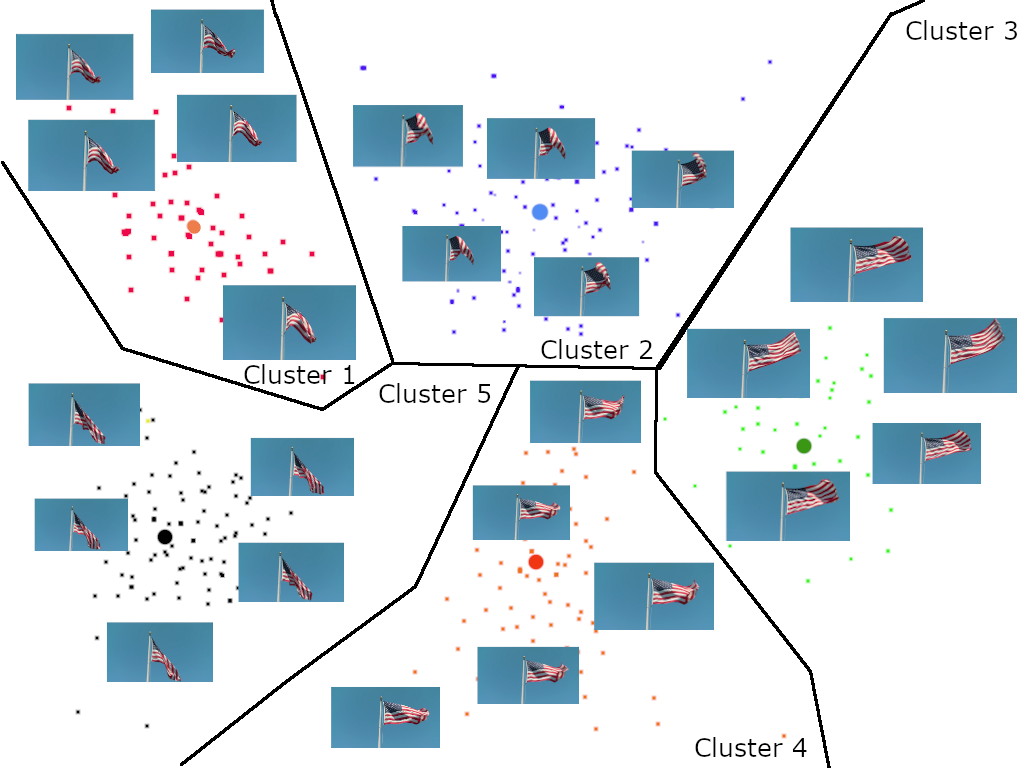}
  \caption{Example of 5-means clustering with Original Frames}
  \label{fig:6}
\end{figure}

Besides SSD, we also used wavelet based K-means clustering, the 5 clusters for the flag clip are shown in Figure \ref{fig:6} along with the uncompressed version of few frames representing that cluster.

\par
\begin{figure}[!h]
  \centering
  \includegraphics[width=\linewidth]{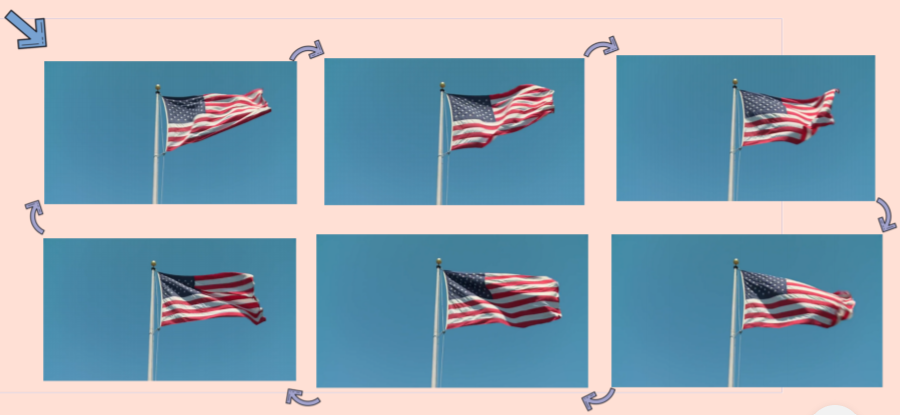}
  \caption{Example Transition of few Frames from the Generated GIF}
  \label{fig:7}
\end{figure}

Figure \ref{fig:7} shows few frames in sequence of the generated loop.

%\begin{table}
%\caption{Example table}
%\centering
%\begin{tabular}{llr}
%\toprule
%\multicolumn{2}{c}{Name} \\
%\cmidrule(r){1-2}
%First name & Last Name & Grade \\
%\midrule
%John & Doe & $7.5$ \\
%Richard & Miles & $2$ \\
%\bottomrule
%\end{tabular}
%\end{table}

%------------------------------------------------

\section{Discussion}
\subsection{Other Video Clips}
\begin{figure}[!h]
  \centering
  \includegraphics[width=0.5\linewidth]{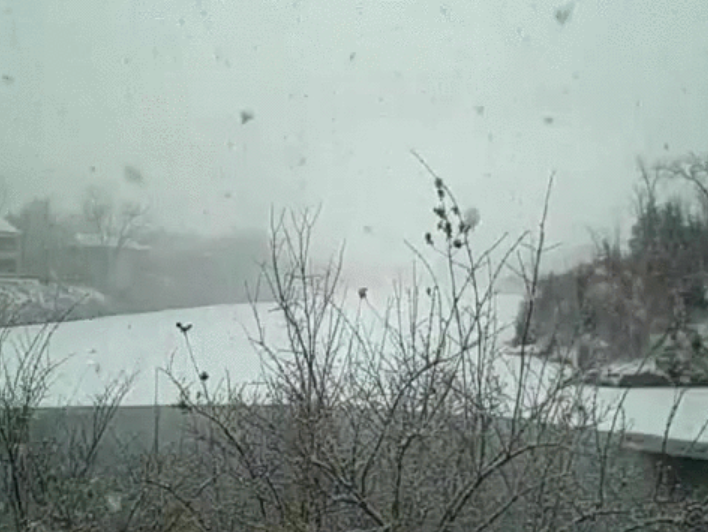}
  \caption{A still from snowfall GIF}
  \label{fig:8}
\end{figure}

Snowfall makes for a great video texture as it's very hard to perceive where the loop happens. Snowflakes all pretty much look the same, so the frame transitions are very smooth. Furthermore, the snow is falling quickly which makes it harder to notice any poorer transitions that may be made. See Figure \ref{fig:8}

\begin{figure}[!h]
  \centering
  \includegraphics[width=0.5\linewidth]{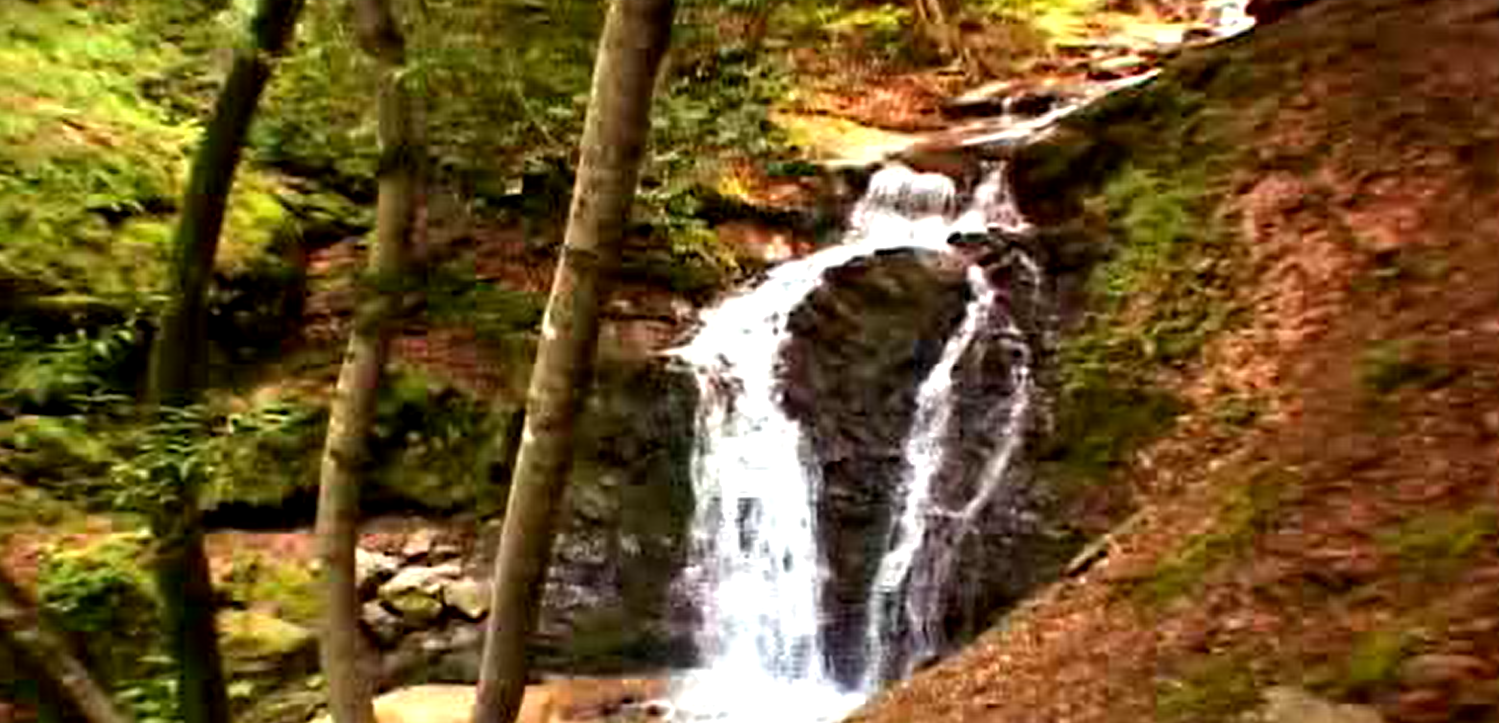}
  \caption{A still from waterfall GIF}
  \label{fig:9}
\end{figure}

The waterfall performs much better with our cross-fading addition, than it did with the original base algorithm. Cross-fading works since water doesn't have a sharp edges that would cause the fade stand out. See Figure \ref{fig:9}

For timelapses, our normalization of light intensity method actually produced smooth results, we can create Cinemagraphs out of this method \cite{cine}, which are high quality GIFs that are smoothly looped, but they were manually made till now and didn't require well-defined patterns.
\begin{figure}[!h]
  \centering
  \includegraphics[width=0.5\linewidth]{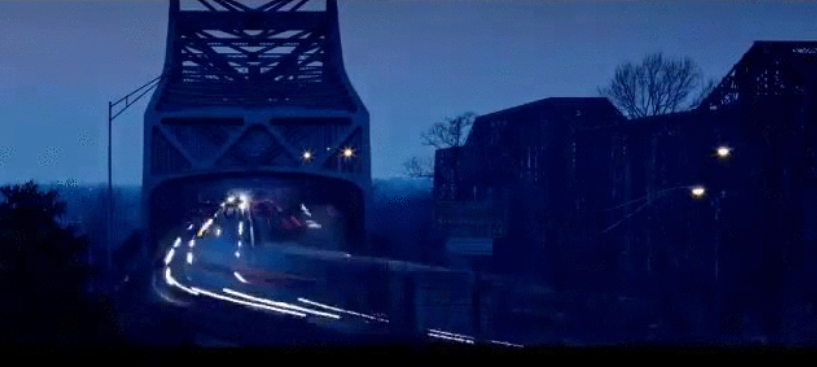}
  \caption{A still from city timelapse GIF}
  \label{fig:10}
\end{figure}

\subsection{Future Work}
Our algorithm only takes into account local neighbors and they are only tried and tested on relatively short video clips (\~2 minutes). For a longer clip, global neighborhood method might require lot of computational resources.\par
Also, many clips might not have a repeatable pattern at all. And thus, our algorithm might not get enough material to create a smooth seamless GIF, resulting into jitters and jumps and looping around a very small portion of the clip.
%----------------------------------------------------------------------------------------
%	REFERENCE LIST
%----------------------------------------------------------------------------------------

\bibliography{ref}{}
\bibliographystyle{ieeetr}

%----------------------------------------------------------------------------------------

\end{document}